\documentclass{article}
\PassOptionsToPackage{numbers, compress}{natbib}

\usepackage[preprint]{neurips_2021}
\usepackage{floatrow}
\newfloatcommand{capbtabbox}{table}[][\FBwidth]
\newcommand{\approach}{\textsc{PegNet}}

\usepackage[utf8]{inputenc} 
\usepackage[T1]{fontenc}    
\usepackage{hyperref}
\usepackage{url}            
\usepackage{booktabs}       
\usepackage{amsfonts}       
\usepackage{nicefrac}       
\usepackage{microtype}      
\usepackage{xcolor}         

\usepackage{amsmath,amsfonts,bm}

\def\eqref#1{equation~\ref{#1}}

\def\1{\bm{1}}

\def\rva{{\mathbf{a}}}
\def\rvb{{\mathbf{b}}}
\def\rvc{{\mathbf{c}}}

\def\rvg{{\mathbf{g}}}
\def\rvh{{\mathbf{h}}}

\def\rvo{{\mathbf{o}}}
\def\rvp{{\mathbf{p}}}
\def\rvq{{\mathbf{q}}}

\def\rvs{{\mathbf{s}}}

\def\rvx{{\mathbf{x}}}
\def\rvy{{\mathbf{y}}}

\def\rmW{{\mathbf{W}}}

\DeclareMathAlphabet{\mathsfit}{\encodingdefault}{\sfdefault}{m}{sl}
\SetMathAlphabet{\mathsfit}{bold}{\encodingdefault}{\sfdefault}{bx}{n}

\usepackage{times}
\usepackage{latexsym}
\usepackage{url}
\usepackage[utf8]{inputenc}
\usepackage{amsmath}
\usepackage{amssymb}
\usepackage{helvet}
\usepackage{courier}
\usepackage{graphicx}
\usepackage{amsfonts}
\usepackage{xspace}
\usepackage{verbatim}
\usepackage{CJKutf8}
\usepackage{multirow}
\usepackage{tabularx}

\usepackage{array}
\usepackage{adjustbox}
\usepackage{subcaption}
\usepackage{booktabs}
\usepackage{algorithm}
\usepackage{comment}
\usepackage{enumitem}
\usepackage{wrapfig}
\usepackage{capt-of}
\usepackage{tikz}
\usepackage{pgfplots}
\usepackage{array}
\usepackage{xcolor,soul}
\usepackage{mathtools}

\usepackage{algorithmic}

\usepackage{color}
\usepackage{colortbl}

\definecolor{mycmark}{RGB}{181, 23, 0}
\definecolor{myxmark}{RGB}{0, 118, 186}
\definecolor{Gray}{gray}{0.9}

\usepackage{microtype}
\title{You Can Do Better!\\ If You Elaborate the Reason When Making Prediction}

\author{Dongfang Li$^1$, Jingcong Tao$^1$,  Qingcai Chen$^{1,2}$, Baotian Hu$^{1}$\\
$^1$Harbin Institute of Technology (Shenzhen), Shenzhen, China \\
$^2$Peng Cheng Laboratory, Shenzhen, China\\
\texttt{crazyofapple@gmail.com, 20s051009@stu.hit.edu.cn}\\
\texttt{qingcai.chen@hit.edu.cn, hubaotian@hit.edu.cn}}

\begin{document}

\maketitle

\begin{abstract}
  Neural predictive models have achieved remarkable performance improvements in various natural language processing tasks. However, most neural predictive models suffer from the lack of explainability of predictions, limiting their practical utility. This paper proposes a neural predictive approach to make a prediction and generate its corresponding explanation simultaneously. It leverages the knowledge entailed in explanations as an additional distillation signal for more efficient learning.  
We conduct a preliminary study on Chinese medical multiple-choice question answering, English natural language inference and commonsense question answering tasks. The experimental results show that the proposed approach can generate reasonable explanations for its predictions even with a small-scale training corpus. The proposed method also achieves improved prediction accuracy on three datasets, which indicates that making predictions can benefit from generating the explanation in the decision process. 
\end{abstract}
\section{Introduction}
\label{sec:introduction}

Neural predictive models (NPMs) have demonstrated their superior ability on various challenging natural language processing (NLP) tasks, especially when implemented  with pre-trained language models (PLM)~\cite{bert,roberta,albert}. However, NPMs are more like black boxes compared to traditional machine learning methods such as logistic regression and decision tree~\cite{samek2017explainable,rudin2019stop}. It is notoriously difficult to understand why NPMs make particular predictions, which significantly limits their practical utility—taking medical scenarios as examples, knowing why NPMs make the prediction is at least as important as the prediction itself~\cite{holzinger2017we,tjoa2020survey}.

Recalling that our human beings brain is also a black box, we almost always make predictions and actions ``\textit{understandable}'' or ``\textit{explainable}'' by explaining the reason. 
Inspired by this observation, there are increasing interests in rationalizing predictions. For example, some works are proposed to generate rationales that support outputs for tasks such as question answering~\cite{rajani-etal-2019-explain, Latcinnik2020ExplainingQA} and natural language inference~\cite{camburu_e_snli_2019,LIREx}. These extracted or generated explanations can partly reflect the decision-making process of the model~\cite{lei16,MelisJ18,Lipton18,Jacovi2020TowardsFI}. 
As extractive rationales are only selected as sub-sequences of the input~\cite{thorne2019generating,deyoung2019eraser}, we focus on generative rationale~\cite{liu2018towards,kumar-talukdar-2020-nile} to provide the justifications for model decisions, which is more consistent with how humans would conduct reasoning from text~\cite{measuring}.

Intuitively, humans who explain examples to themselves learn better, make more accurate self-assessments of their understanding, and use analogies more appropriately while solving problems~\cite{chi1989self,vanlehn1992model,chi1994eliciting}. 
To generate faithful and plausible explanations, we similarly consider predicting the label and generating the corresponding explanation jointly. Nevertheless, this line of research is likely to be sensitive to hyperparameters and requires the complicated training process~\cite{Jain2020LearningTF}. For example, previous works simulate the intractable sampling step by proposing optimization procedures based on reinforcement learning approaches~\cite{lei16,leakage} and reparameterization techniques~\cite{Latcinnik2020ExplainingQA,bastings2019interpretable}. Instead, our approach leverages the knowledge entailed in explanations as an additional distillation signal for more efficient learning.  

In this paper, we propose a neural predictive approach named~\approach{} coupled with powerful pre-trained language  models  to  make  the prediction and generate its corresponding explanation simultaneously. The proposed~\approach{} consists of a label predictor and an explanation generator. They are jointly trained in the multi-task learning paradigm. The label predictor emits the final prediction, while the generator is used to elaborate the explanation. Since the classifier only relies on the generator when training, it prevents information leakage~\cite{leakage} between explanations and labels. And we use the weighted loss function to connect these two different components. 
We evaluate the proposed method on three tasks: Chinese medical Multiple-Choice Question Answering (MCQA), English Natural Language Inference (NLI), and commonsense question answering. As the lack of interpretability is a key factor that limits wider adoption of NPMs in the medical domain, we provide more analysis for the medical MCQA task.

The contributions of this paper are summarized as follows: 
\begin{itemize}[leftmargin=*]
	\item We propose a ``\textit{predict-and-explain}'' end-to-end approach (\approach{}) for natural language reasoning tasks where explainability plays a pivotal role already during learning. It takes advantage of discriminative models to predict labels while condensing the information in the explanations to make them more related to labels. Besides, we introduce a quantitative method to assess the quality of explanations. 
    \item Experiment results show that the performance on these tasks improves while generating fluent and informative natural language explanations. For the MCQA task, we consider Chinese medical exams as the source of questions and collect human-written natural language explanations for each question. Human evaluations from different criteria (e.g., causality) further demonstrate that our method outperforms baselines.
\end{itemize}

\section{Related Work}

\paragraph{Interpretability of NLP Models} 
Since deep learning became the dominant paradigm in NLP research, how to interpret the predictions of neural models has become an essential part of model transparency. One approach is to study the feature attributions of the model by extracting salient parts from the inputs~\cite{LIME,anchors,shap}. Recently, attention mechanism~\cite{bahdanau2014neural} is gaining popularity in NLP models, which facilitates understanding of the model through attention weight visualization~\cite{Jain2020LearningTF}. However, it remains uncertain whether attention weights can provide reliable insights for the decision process of the model~\cite{att2,wiegreffe-pinter-2019-attention,pruthi2020learning}. An alternative approach is to generate natural language explanations for the model's decisions. It is typically done by training the model on free-text natural language explanations; however, these explanations are expensive to collect and not uniquely fixed for each answer.
According to the way of generating explanations and making predictions, previous works can be further divided into four types: 1) generating explanations for each label first, then using the predicted label to choose the corresponding explanation~\cite{LIREx,kumar-talukdar-2020-nile}; 2) predicting the label first, then generating the explanation with the label and original inputs~\cite{camburu_e_snli_2019}; 3) generating explanations first and then using the generated explanations to predict the labels~\cite{lei16,Jain2020LearningTF};  4) predicting the label and generating the corresponding explanation jointly~\cite{Latcinnik2020ExplainingQA,zhou2020nips,melis2018towards,Narang2020WT5TT}.
We collect a medical MCQA dataset with natural language explanations corresponding to each question, which is more specialized and will serve as a testbed in the medical domain.

\paragraph{Explanation Generation Methods} 
Our work aims to output natural language explanations together with labels. Different from ante-hoc or post-hoc methods~\cite{camburu_e_snli_2019,kumar-talukdar-2020-nile,LIME}, our model focuses on the performance of the inference task while encouraging the generated explanations to be faithful to the predictions. 
There are also some works focusing on generating natural language explanations~\cite{zhou2020nips,prasad2020extent,carton2020evaluating,jacovi2020aligning}. For example, ~\citet{Narang2020WT5TT} leverage the text-to-text framework~\cite{raffel2019t5} to train language models to output natural language explanations along with their prediction. ~\citet{Latcinnik2020ExplainingQA} propose an explainable PLM-based model for MCQA by using XLNet~\cite{yang2019xlnet} for both generation and classification.  Besides, there are some recent works~\cite{leakage,pruthi2020evaluating} focused on evaluation metrics for explanations such as the measurement of label-rationale association. 

\section{Method}
\begin{figure*}[hpt]
\centering
%
\includegraphics[scale=1.0,width=0.9\linewidth]{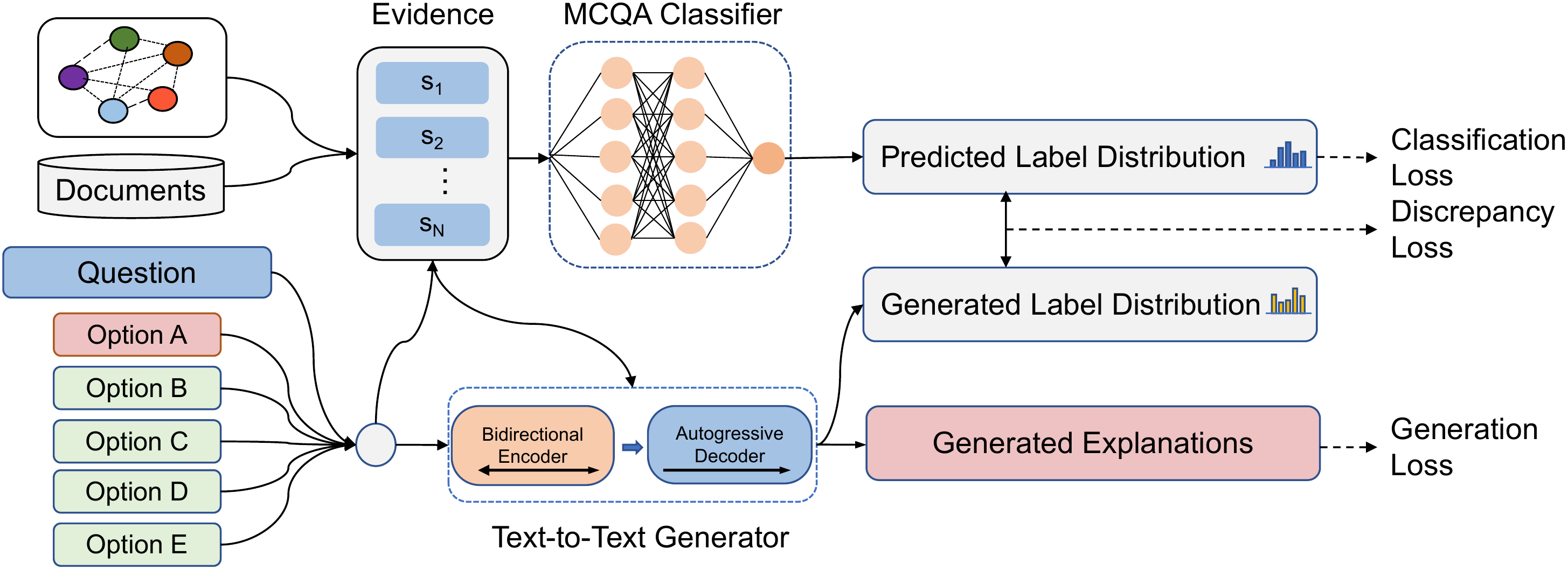}
\caption{Overall architecture of the \approach{} for multiple-choice question answering. We first collect evidence for each question-option pair from heterogeneous sources. Then for generation, question-all-options text and retrieved evidence by using question only (refer as question context) are concatenated to send into the text generator, and each question-option pair is fed into output layer for classification. Finally, we use the discrepancy loss to connect the two components.}
\label{fig:model}
\end{figure*}
We first describe the overall framework of~\approach{}, followed by the basic components we use. Then, we introduce the details of our proposed approach, including how to integrate the classifier and generator, and the training method when only a portion of samples are provided with human-written explanations.
\subsection{Overall Framework}



We work in the two following settings:  multiple-choice question answering (MCQA) and natural language inference (NLI). The former is to answer multiple-choice questions that rely on domain-specific knowledge or commonsense reasoning, and the latter is to determine the relationship between premise-hypothesis pairs, i.e., entailment, contradiction or no relationship (neutral). 

The overall framework consists of two basic components: the text-to-text generator and the classifier, whose parameters are independent, as shown in Figure~\ref{fig:model}.  For the MCQA task, the input to the pre-trained autoregressive generator is a combination of questions with all options: ``\{question\} The options are \{option$_1$\}, \{option$_2$\}...''. The target output is ``My commonsense tells me that \{explanation\}''. We further retrieve evidence related to the question, referred to as the question context. On the other hand, the classifier is a pre-trained language model with a multiple-choice classification head on the top layer, whose input is the question plus each answer candidate with special tokens delimiting them (e.g., ``[CLS] \{question\}  \{option$_j$\} [SEP] \{evidence$_j$\} [EOS]''). For NLI, we similarly perform the three-way classification to predict the relation of premise-hypothesis, while the input to generator is changed to a combination of premise-hypothesis pairs (i.e., ``nli \{premise\} \{hypothesis\}'').
The generator training is guided by humans' explanations, while training targets of the classifier are the answers to questions. Furthermore, we connect these two components through loss function. After training, we take the classifier's output as the predicted label and use the generator's output as generated explanations. We also use the beam search to decode multiple generated candidates. To discourage repetitions in the generated text, we penalize the scores of previous generated tokens~\cite{keskar2019ctrl}.

\subsection{Basic Components}
For classification, we use pre-trained language models trained with masked language modeling (MLM) loss. Specially, we apply ALBERT~\cite{albert} and RoBERTa model~\cite{roberta,yiming19} in English and Chinese, respectively. For the downstream classification task, a prediction head is added to the hidden state of the first token in the last layer.

For generation, we use the large pre-trained text-to-text transformer named T5~\cite{raffel2019t5} and its multilingual version, mT5~\cite{xue2020mt5}, as the generator. The pre-training of T5 contains both unsupervised and supervised parts. The unsupervised part uses a corpus of nearly 800G (called C4 in the paper), while the training objective is similar to BERT~\cite{bert}, except that it is changed to a Seq2Seq version. It can be considered as a variant of MLM task. On the other hand, the supervised part collects data from common NLP tasks and uniformly transforms them into Seq2Seq tasks for training. For example, for sentiment classification, the input is ``sentiment: I feel good today'', and the output is ``positive''. 

\subsection{Predicting and Rationalizing}
For simplicity, we describe our approach under the MCQA setting.
Given $N$ input samples $\mathcal{D} = \{(\rvq_{i}, \{\rvo_{i}^{j}\}, \rva_{i}, \rvs_{i})\}_{i=1}^N$, we concatenate question $\rvq_{i}$, each answer candidate $\{\rvo_{i}^{j}\}$, and the evidence $\rvs_{i}$ (if have) by the specific separator tokens.
For each candidate $\rvo_{i}^{j}$, we then compute a score $p_i^j$ by passing the hidden states of the first token $\rvh_{i}^j \in \mathbb{R}^{d}$ through an prediction head: 
\begin{equation}
\small
p_{i}^j = \rmW_2\tanh( \rmW_1 \rvh_{i}^j + \rvb_1),
\end{equation}
\begin{equation}
\small
\mathcal{L}_{ce} = -\sum_{i=1}^N\rva_{i}\log Softmax(\rvp_{i}),
\end{equation}
where $\rmW_1\in \mathbb{R}^{d\times d}$, $\rvb_1 \in \mathbb{R}^d$, $\rmW_2 \in \mathbb{R}^{1\times d}$, and $d$ denotes the hidden size.
Finally, we normalize scores across all options for a given question via the softmax layer.
The option with the highest probability is chosen as the predicted label.

On the other hand, text generation with a text-to-text model is to generate an output text sequence $\rvy_{i} = [y_{i}^1, \ldots, y_{i}^T]$ with length $T$ conditioned on the input text sequence $\rvx_{i} = [x_{i}^1, \ldots, x_{i}^M]$ with length $M$. A typical approach to the text generation is to leverage the encoder-decoder architecture to parameterize the conditional distribution. We maximize the conditional log likelihood $\log l(\rvy| \rvx)$ for a given sample $(\rvx_{i}, \rvy_{i})$ as follows:
\begin{equation}
\small
    \rvh_{i}^t = Decoder(y_{i}^{t-1}, \mathbf{c}_{i}), \mathbf{c}_{i} = Encoder(\rvx_{i}),
\end{equation}  
\begin{equation}
\small
    l(y_{i}^t | \rvy_{i}^{<t}, \rvx_{i}) = \text{Softmax}(\rmW\rvh_{i}^t + \rvb_2),
\end{equation}
\begin{equation}
\small
    l(y_{i}^1, \ldots, y_{i}^T | \rvx_{i}) = \prod_{t=1}^T l(y_{i}^t |\rvy_{i}^{<t}, \rvx_{i}),
\end{equation}
\begin{equation}
\small
    \mathcal{L}_{mle} = -\sum_{i=1}^N\log l(\rvy_{i}|\rvx_{i}),
\end{equation}
where $\mathbf{C}_{i} =[\rvc_{i}^1 \cdots \rvc_{i}^M] \in \mathbb{R}^{d\times M}$ denotes the concatenation of the hidden states of the source tokens $\rvx_{i}$ and $\mathbf{H}_{i} =[\rvh_{i}^1 \cdots \rvh_{i}^T] \in \mathbb{R}^{d\times T}$ denotes the concatenation of the hidden states of the generated tokens $\rvy_{i}$ respectively. 

To encourage the generator to produce consistent outputs that coordinate with the classifier, the hidden states of the decoder is firstly fed into a fully connected layer. Then we perform max-pooling and softmax operations to get the generated label distribution with the same dimensions as the number of options. Then we compute the cross-entropy loss for this distribution:
\begin{equation}
\small
    \rvg_{i} = Max\-pooling(\rmW_3 \mathbf{H}_{i} + \rvb_3),
\end{equation}
\begin{equation}
\small
    \mathcal{L}_{ce_g} =  -\sum_{i=1}^N\rva_{i}\log Softmax(\rvg_{i}).
\end{equation}
Finally, we use the discrepancy loss to connect the two components. Compared to the generator, the classifier uses multiple inputs for each question and thus learning much slower. 
Hence, we apply an additional knowledge distillation loss, specifically Kullback-Leibler divergence loss with temperature equals $1.0$, between the logits of the classifier $\rvp_i$ and the logits of generator model $\rvg_i$ to accelerate the learning, as shown below:
\begin{equation}
\small
\mathcal{L}_{dis} = \frac{1}{N} \sum\limits_{i=1}^{N} D_{KL}(\log Softmax(\frac{\rvp_i}{\tau})\ ||\ Softmax(\frac{\rvg_i}{\tau})),
\end{equation}
where the $\tau$ denotes the temperature term. The overall training objective is divided into four parts: 
\begin{equation}
\small
\mathcal{L} = \lambda_1\mathcal{L}_{ce} + \lambda_2\mathcal{L}_{mle} + \lambda_3\mathcal{L}_{ce_g}  + \lambda_4\mathcal{L}_{dis},
\end{equation}
where $\lambda_1, \lambda_2, \lambda_3, \lambda_4$ are the weight parameters of each part.

When only part of samples have human-written explanations, we apply different input formats of the generator for consistency training. For samples without annotated explanations, the input format remains unchanged while their target format becomes ``The answer is \{golden label\}''. For other samples with golden explanations, we prefix the input format with the word ``explanation'', and the target format changes to ``The answer is \{golden label\}. My commonsense tells...'' accordingly.

\section{Experiment}

\subsection{Datasets}
\label{sec:dataset}
For the MCQA task, we collect Chinese medical multiple-choice questions following~\citet{li2020towards}. Each sample contains one question, five options, and question-option evidence retrieved from medical books and medical knowledge graphs. There is only one correct answer for each question, and the source of the data is mainly from medical qualification examinations. After preprocessing the data, medical experts are asked to analyze that why this option is selected as the correct answer as explanations. To control the quality of open-ended annotations, we double-check to ensure that the correct answer can be obtained by reasoning given questions, options and explanations (rather than through trivial patterns). Then we shuffle the data and divide it into the training set, development set, and test set as Chinese Medical Explanations (CME) dataset. The distribution of correct answers among all options is balanced, and we summarize this dataset statistics in Table~\ref{tab:data_stat}. We further collect 17,629 samples without human-written explanations. 

\begin{table*}[ht]
\centering
\small
\begin{tabular}{lccc}
\toprule
\textbf{} & \textbf{Train}  & \textbf{Dev} & \textbf{Test} \\
\midrule
\# Questions & $2,172$ & $240$ & $242$ \\
Avg. words of questions & $21.79$ & $25.21$ & $20.66$ \\
Avg. words of options & $4.18$ & $4.24$ & $4.63$ \\
Avg. words of explanations & $58.23$ & $51.55$ & $59.59$ \\
\midrule
Candidate answer per problem & \multicolumn{3}{c}{$5$} \\
\bottomrule
\end{tabular}
\caption{Statistics of our CME dataset. The explanations are generated by human experts, which is highly specialized and implies the expertise to be examined. It contains the reasons for choosing the correct answer and why the other options are wrong (contrastive part).}
\label{tab:data_stat}
\end{table*}

To evaluate the proposed approach, we also use two public datasets e-SNLI~\cite{camburu_e_snli_2019} and CoS-E~\cite{rajani-etal-2019-explain}. 
~\citet{camburu_e_snli_2019} recently crowd-sourced the e-SNLI data set by adding annotations to each data instance from SNLI~\cite{bowman2015large}, which is based on the golden label of each sample and provides natural language explanations marked by the annotator. The training set, development set, and test set consist of 532,012, 9,842, and 9,824 samples, respectively. ~\citet{rajani-etal-2019-explain} created the CoS-E data set with free-text explanations for commonsense MCQA task. We used two versions of CoS-E (i.e., v1.0, v1.11) without private test set because the last version has noise in its annotations~\cite{measuring,Narang2020WT5TT}. Its training set and the development set consist of 7,610/9,741 and 950/1,221 multiple-choice questions. 
\subsection{Training Details}
\label{sec:implement}
For classification, we use ALBERT-xxlarge-v2 model~\cite{albert} and the Chinese Roberta-large model~\cite{yiming19}. For generation, we use the base version of T5 with 220M parameters and its multilingual version with 600M parameters. We use AdaFactor optimizer with a learning rate of $0.001$ and AdamW optimizer with a learning rate of $2 \times 10^{-5}$ for model training, respectively. The maximum sequence length, the learning rate warmup proportion and the training epoch are set to $256$, $0.1$, and $10$.  We use gradient clipping to a maximum norm of $1.0$ and a dropout rate of $0.1$. We use beam search (20 beams) until an end token is generated (or for a max length of 200 tokens). The repetition penalty is $1.5$. 
Our models are trained with 32GB NVIDIA Tesla V100s. We select the model with the best accuracy on the development set.
\subsection{Evaluation Metrics}
\label{sec:evaluation}
For classification, we report the accuracy of each method.  For the MCQA task, we further adopt another metric \textit{Accuracy$_{y(e)}$}. It is a quantitative measurement by considering the mean accuracy score of trained MCQA models, which are learned from original question-option pairs and their evidence. 
When testing, we remove the question and replace the evidence with generated explanations or ground-truth natural language explanations. 
The basic idea is to test whether only using the generated explanation can provide enough information for predicting the answer
~\cite{deyoung2019eraser, measuring}. However, the problem is about how much information the prediction relies on generated explanations. We mitigate this problem as all models are fair when testing, and there are no human-written explanations added during training. 
Moreover, as \citet{measuring} suggested that the quality of explanations generated by the joint method needs to be verified further, we conduct human evaluations to evaluate their quality.

\subsection{Experiment Results}
\begin{table*}[ht]
    \small
    \begin{tabular}{lccc}
    \toprule
         \multirow{1}{*}{\textbf{Model}} & \multicolumn{1}{c}{\textbf{e-SNLI}} & \multicolumn{1}{c}{\textbf{CoS-E v1.11 / v1.0}} & \multicolumn{1}{c}{\textbf{CME}}  \\
        \midrule
        Human         &$98.8$\textsuperscript{$\mathsection$}    & $80.1$\textsuperscript{$\mathsection$} / $90.5$\textsuperscript{$\mathsection$}  &  $87.6$\textsuperscript{\enspace}  \\ \midrule
        DIIN~\cite{DBLP:conf/iclr/GongLZ18} & $88.9$ & - & - \\
        ESIM~\cite{DBLP:conf/acl/ChenZLWJI17} & $88.6$ & - & - \\
        DRCN~\cite{DBLP:conf/aaai/KimKK19} & $90.1$ & - & - \\
        MT-DNN~\cite{DBLP:conf/acl/LiuHCG19} & $91.6$ & - & - \\
        CoS-E-open-ended~\cite{rajani-etal-2019-explain} & - & $58.2$ / $65.5$ & - \\
        CAGE-reasoning~\cite{rajani-etal-2019-explain} & - &  $55.7$ / $72.6$ & - \\
        KMQA~\cite{li2020towards} & - & - & $54.4$ \\
        BERT~\cite{bert}      & $90.8$      & $56.7$ / $63.8$        & $41.5$    \\
        WT5-base~\cite{Narang2020WT5TT}      & $90.9$      & $59.4$ / $66.1$       & $43.8$   \\
        WT5-large~\cite{Narang2020WT5TT}  & $92.3$ & $\bf82.7$ / $80.6$ & $56.4$ \\ 
       \approach{} (ours) & $\bf92.5$\textsuperscript{\enspace} & $80.8$\textsuperscript{\enspace} / $\bf87.0$\textsuperscript{\enspace}	 & $\bf62.8$\textsuperscript{\enspace}  \\
        \bottomrule
        
    \end{tabular}
    \caption{The results of~\approach{} and several baselines on the data set we studied. 
    The human results are from previous works: \textsuperscript{$\mathsection$} \citet{measuring}.
    The base components of ours model are T5-base version.
    }
    \label{tab:main_result}
    
\end{table*}

From Table~\ref{tab:main_result}, our method delivers competitive performance compared to the previous WT5-base method under the condition of T5-base. Additionally, it is slightly better than the previous state-of-the-art model in the SNLI test set. Moreover, it is superior to the model using human-annotated explanations in the CoS-E v1.11 dataset.  
The reason why jointly training is better than models training with human explanations is due to that the CoS-E dataset (even v1.0) includes many meaningless and uninformative explanations, such as ``I consider it the best option because they have to be more true''. 
For the CME dataset,~\approach{} also achieves improved performance compared to strong competitive methods.
\subsection{Human Evaluation}
\begin{table*}[t]
    \centering
    \small
    
    \resizebox{1.0\columnwidth}{!}{
    \begin{tabular}{l  ccc ccc ccc ccc}
    \toprule
        \multirow{2}*{\textbf{Method}} & \multicolumn{3}{c}{\textbf{Fluency}} & \multicolumn{3}{c}{\textbf{Causality}} & \multicolumn{3}{c}{\textbf{Informativeness}} & \multicolumn{3}{c}{\textbf{Repetition}} \\
         & Win & Lose & Tie & Win & Lose & Tie & Win & Lose & Tie & Win & Lose & Tie \\
         \midrule
       ~\approach{} vs. Human & $0.04$ & $0.22$ & $0.74$ & $0.14$ & $0.50$ & $0.36$ & $0.02$ & $0.60$ & $0.38$ & $0.00$ & $0.18$ & $0.82$\\
       ~\approach{} vs. Only Generation & $0.18$ & $0.12$ & $0.70$ & $0.24$ & $0.14$ & $0.62$ & $0.20$ & $0.16$ & $0.64$ & $0.08$ & $0.16$ & $0.76$ \\
       ~\approach{} vs. WT5-base & $0.18$ & $0.12$ & $0.70$ & $0.40$ & $0.08$ & $0.52$ & $0.32$ & $0.12$ & $0.56$ & $0.14$ & $0.14$ & $0.72$ \\
        \midrule
        Cohen's kappa coefficient & \multicolumn{3}{c}{$0.4367$} & \multicolumn{3}{c}{$0.4517$} & \multicolumn{3}{c}{$0.3892$} & \multicolumn3{c}{$0.3340$}\\ 
        \bottomrule
    \end{tabular}
    }
    \caption{Human evaluation results of our method compared to other baselines. The scores are the percentages that our method wins/ loses/ties in pair-wise comparison.}
    \label{tab:human}

\end{table*}
To assess the quality of the generated explanations, we perform manual evaluations for our method on the CME dataset. Two postgraduate students 
from top-tier medical school are recruited as annotators for pairwise comparisons. Each annotator is given 50 random samples in the test set.
Each of these samples includes the question, all options, the golden answer, the predicted answer generated by two different methods, and the corresponding explanations.
Each sample is evaluated on four metrics: fluency, causality, informativeness and repetition. Each annotator is required to give a preference among ``win'', ``tie'' and ``lose''. 

The definitions of these metrics are: (1) \textit{Fluency} is designed to measure whether the texts are fluent. It is used to measure the grammatical correctness and readability of the explanation; (2) \textit{Causality} (i.e., faithfulness) represents the degree of association (in terms of reasonability) between the predicted answer and the generated explanation. It is used to measure whether the explanation accurately describes the true machinery of the model’s prediction; (3) \textit{Informativeness} represents the amount of information of the texts itself. It is used to measure the extent to which the new information conveyed in the explanation is helpful to answer the question; (4) As the pre-trained generative model approach is prone to generating sentences that contain many meaningless repetitions, we use the \textit{Repetition} metric. Compared to fluency, it is a subjective measure of whether the sentence contains less nonsensical repeated fragments. 

The results are shown in Table~\ref{tab:human}, where our method outperforms the baselines on the first three metrics. Specially, our method is better than the explanation-only generation approach in terms of causality. However, the generated texts are still inferior to the human-written explanations, especially on the metrics of causality, informativeness and repetition.
We compute cohen's kappa score~\cite{cohen1960coefficient} to measure inter-annotator agreement. The calculated scores indicate a moderate agreement among annotators on the first two metrics, while the degree of agreement is fair on the last two.

\subsection{Error Analysis and Ablation Study}
\begin{table*}[ht]
\centering
    \small
    \resizebox{1\columnwidth}{!}{
    \begin{tabular}{lll}
    \toprule
    \textbf{Error Type} & \textbf{Ratio (\%)}  & \textbf{Explanations}\\
    \midrule
    \multirow{2}*\textbf{Inconsistent} & $18.60$ & \begin{CJK}{UTF8}{gbsn}\underline{阿司匹林泡腾片}可能引起胃溃疡...\end{CJK} \\
    &  & \underline{Aspirin effervescent tablets} may cause gastric ulcers... \\
    \multirow{2}*\textbf{Repetition} & $23.26$  & \begin{CJK}{UTF8}{gbsn}可引起子痫发作的药物有...\underline{甲硫酸镁、甲硫酸镁、甲硫酸镁}... \end{CJK} \\
    & &  The drugs that can cause eclampsia are...\underline{methosulfate, methosulfate, methosulfate}... \\
    \multirow{2}*\textbf{Contradiction} & $16.28$  & \begin{CJK}{UTF8}{gbsn}...,从而\underline{降低}血糖代谢速度,\underline{提高}血糖代谢速度...\end{CJK} \\
     & &  ..., thereby \underline{reducing} the rate of blood glucose metabolism, \underline{increasing} the rate of ... \\
    \multirow{2}*\textbf{Others} & $41.86$  & \begin{CJK}{UTF8}{gbsn}\underline{血中胰岛素}和\underline{C肽}水平很低甚至检测不出。 \end{CJK} \\
    & &  \underline{The blood insulin} and \underline{C-peptide} levels are very low or even undetectable. \\
    \bottomrule
    \end{tabular}
    }
    \caption{The percentage of different error types and typical cases of explanation generated by our method. Underlined texts denote key phrases in the generated text of each error type.}
     \label{tab:error}
\end{table*}
\begin{table*}[ht]
\small
\centering
\begin{tabular}{lccc}
\toprule
\textbf{Model} & \textbf{Accuracy}   & \textbf{BLEU}  & \textbf{Accuracy$_{y(e)}$} \\
\midrule
\approach{} & $62.81$ & $23.66$ & $39.67$ \\
WT5-base~\cite{Narang2020WT5TT} & $43.80$ & $27.03$ & $39.26$ \\
\midrule
\textsc{Classifier} w/ only QA pairs (i.e., \textit{y}$|$\textit{x}) & $28.92$ & - & - \\
\approach{} w/o discrepancy loss (i.e., \textit{y}$|$(\textit{x}, \textit{evidence})) & $58.68$ & - & $38.01$ \\
\approach{} w/o question context & $62.39$ & $19.59$ & $38.84$\\ \midrule
\textsc{Classifier} w/ evidence \& no-expl-samples & $69.01$& -   & - \\
\approach{} w/ no-expl-samples & $70.66$ & $22.02$ & $51.23$ \\ 
\textsc{Classifier} w/ expls (i.e., \textit{y}$|$(\textit{x}, \textit{expls})) & $87.60$& -   & $66.12$ \\
\bottomrule
\end{tabular}
\caption{Ablation study in the CME dataset. Here \textit{y}$|$\textit{x} means predicting labels by question-option pairs in training and test phase,  \textit{evidence} means using retrieved evidence and \textit{expls} means using golden explanations. }
\label{tab::ablation}
\end{table*}
\begin{table*}[!ht]
\centering
\small
\begin{tabular}{p{\textwidth}}
\toprule
\textbf{The \textit{y}$|$\textit{x} classifier inputs}:\\
for i$_{th}$ option: $[CLS] \{question\} [SEP] \{option_i\} [EOS]$ \\
\midrule
\textbf{The \textit{y}$|$(\textit{x}, \textit{evidence}) classifier inputs}:\\
for i$_{th}$ option: $[CLS] \{question\} \{option_i\} [SEP] \{evidence_i\} [EOS]$ \\
\midrule
\textbf{The \textit{y}$|$(\textit{x}, \textit{expls}) classifier inputs}:\\
for i$_{th}$ option: $[CLS] \{question\} \{option_i\} [SEP] \{golden\_explanations_i\} [EOS]$ \\
\midrule
\textbf{To calculate Accuracy$_{y_e}$ when testing, the classifier inputs}:\\
for i$_{th}$ option: $[CLS] \{option_i\} [SEP] \{explanations_i\} [EOS]$ \\
\midrule
\textbf{The generator inputs is}:\\
$[CLS] \{question\}\ The\ options\ are\ \{option_1\} \{option_2\} ... reference: \{question\_context\} [SEP]$ \\
\midrule
\textbf{The generator inputs without question context}:\\
$[CLS] \{question\}\ The\ options\ are\ \{option_1\} \{option_2\} ... [SEP]$ \\
\midrule
\textbf{The generator outputs}:\\
My commonsense tells me that \{generated explanation\}.\\
\bottomrule
\end{tabular}
\caption{The input/output formats of each variant. Here $[CLS]$, $[SEP]$ and $[EOS]$ mean symbols for segmentation. Note that it depend on different pre-trained language models.}
\vspace{-2mm}
\label{tab:t5_data_format}
\end{table*}

To analyze the types of errors in the explanations produced by the model, we manually examine all cases in which the causality was not as good as the latter in a pairwise comparison between the model and human-written explanations. We annotate four types of errors from the failed explanations: inconsistent, repetition, contradiction, and others (e.g., unrelated).  
As shown in Table~\ref{tab:error}, there still exists inconsistent explanations. For example, in the first line of Table~\ref{tab:error}, the predicted label is another option ``\begin{CJK}{UTF8}{gbsn}左氧氟沙星片\end{CJK} (Levofloxacin Tablets)'', but the explanation is about the generic description of  ``\begin{CJK}{UTF8}{gbsn}阿司匹林泡腾片\end{CJK} (Aspirin Effervescent Tablets)'', which is related to the correct answer to the question. It means that the generated text may be biased toward correct label. This phenomenon has been observed by previous works~\cite{LIREx,kumar-talukdar-2020-nile}. 
The second type of error is repetition, which is a problem we encountered during the decoding of the pre-trained generative model~\cite{rept1,rept2}. The last two error types indicate that the generated text itself contains some contradictions and noise. 
Actually, we observe that some plausible explanations (even hallucinations) are reasonable but increase the difficulty of reading. It shows that it is challenging for the generator to generate ideal support evidences for the prediction. 

As shown in Table~\ref{tab::ablation}, we conduct the ablation experiments on the CME test data by masking the input. For comparison, we include baseline without the supervision of explanations. We provide the input-output formats of each variant in Table~\ref{tab:t5_data_format}. The model without discrepancy loss decreases the accuracy by 4.13\%. It means joint learning not only provides a window to describe system internals in an understandable way, but also leads to improvement. We utilize the BLEU score \citep{papineni2002bleu} to compare a predicted explanation against the ground-truth explanation. The BLEU value decreases when trained together with the samples without annotated explanations, but the other two metrics increase. We found that the accuracy scores correlate with the proposed metric \textit{Accuracy$_{y(e)}$} for automatic explanation evaluation. It implies that researchers could find it useful to augment standard measures by taking fidelity into account.

\begin{table*}[h]
    \centering
    \small
    \resizebox{1\columnwidth}{!}{
    \begin{tabular}{|p{1\columnwidth}|}
        \hline
        \textbf{Question:} \begin{CJK}{UTF8}{gbsn}下列关于甲状腺功能亢进症患者教育的说法，错误的是哪一个？\end{CJK} Which of the following statements about patients with hyperthyroidism is wrong? \\
        \hline
        \textbf{Options:} \begin{CJK}{UTF8}{gbsn} A.禁食富含碘的食物; B.避免服用含碘的药物; C.哺乳期伴甲状腺功能亢进症患者首选甲巯咪唑; $\checkmark$ D.避免饮用咖啡等兴奋性饮料; E.给予充足的热量、蛋白质和维生素。\end{CJK} A. Abstain from iodine-rich foods; B. Avoid iodine-containing drugs. C. Thiamazole is preferred in patients with hyperthyroidism during lactation; $\checkmark$ D. Avoid excitatory drinks such as coffee; E. Provide adequate calories, protein and vitamins. \\ 
        \hline
        \textbf{Question Context:} \begin{CJK}{UTF8}{gbsn}甲状腺功能亢进症、甲状旁腺功能亢进症、肾上腺皮质功能减退症、肝功能严重障碍等。\end{CJK} Hyperthyroidism, hyperparathyroidism, adrenal hypofunction, severe liver dysfunction, etc. \\
        \hline
        \textbf{Golden Explanation:} \begin{CJK}{UTF8}{gbsn}甲巯咪唑和丙硫氧嘧啶可由乳汁分泌， 引起婴儿甲状腺功能减退，故服药后不宜哺乳；\textcolor{blue}{若必须用药，首选丙硫氧嘧啶，因其乳汁分泌量较小。}\end{CJK} Thiamazole and propylthiouracil can be secreted by breast milk, causing hypothyroidism in infants, so they should not be breastfed after the administration of the medicine; \textcolor{blue}{if medication is necessary, propylthiouracil is preferred because its milk secretion is small.}\\
        \textbf{WT5-base:} \begin{CJK}{UTF8}{gbsn}\textcolor{red}{含碘药物可与碘络合,降低碘吸收,同时可促进碘的吸收。含碘药物可促进胆汁的分泌,延缓胃排空的速度,使碘在体内的吸收显著增加。该药有碘、氢氧化铝凝胶和铝碳酸镁等形成保护膜,使维生素的吸收显著增加。}\end{CJK} \textcolor{red}{Iodine-containing drugs can complex with iodine, reduce iodine absorption, and promote iodine absorption. Iodine-containing drugs can promote the secretion of bile, delay the speed of gastric emptying, and significantly increase the absorption of iodine in the body. The medicine has iodine, aluminum hydroxide gel and hydrotalcite to form a protective film, which significantly increases the absorption of vitamins.} \\
        \textbf{Only Generation (mT5-Base):} \begin{CJK}{UTF8}{gbsn} 甲亢宜釆用最小有效剂量的抗甲状腺药物,因妊娠期用药甲蔬咪哇、丙硫氧嘧啶等可透过胎盘并引起胎儿甲状腺肿大及功能减退,在分娩时造成难产、窒息。由于甲蔬嘧啶有新生儿皮肤缺损的致畸作用报道,\textcolor{red}{妊娠期}\textcolor{blue}{妇女甲亢首选丙硫氧嘧啶。}\end{CJK} Hyperthyroidism should use the smallest effective dose of antithyroid drugs, because medications such as methamphetamine and propylthiouracil during pregnancy can pass through the placenta and cause fetal goiter and hypofunction, which can cause dystocia and suffocation during delivery. Due to reports of teratogenic effects of methamphetamine on neonatal skin defects, \textcolor{blue}{propylthiouracil is the first choice for} \textcolor{red}{pregnant} \textcolor{blue}{ women with hyperthyroidism.} \\
        \textbf{Only Generation (NEZHA-Gen):} \begin{CJK}{UTF8}{gbsn}\textcolor{red}{甲亢治疗的主要不良反应有头晕、头痛、恶心、呕吐、腹泻、便秘、嗜睡、消化道溃疡、胃肠道不适等。}对于妊娠期妇女，可选用口服碘制剂，如果碘摄入过多，会导致碘缺乏，严重者会发生碘中毒，甚至危及生命。\end{CJK} \textcolor{red}{The main adverse reactions of hyperthyroidism treatment include dizziness, headache, nausea, vomiting, diarrhea, constipation, lethargy, digestive ulcer, gastrointestinal discomfort and so on.} For pregnant women, oral iodine preparations can be used. If iodine intake is too much, it will lead to iodine deficiency. In severe cases, iodine poisoning may occur and even life-threatening. \\
        \hline
        \textbf{Ours:} \begin{CJK}{UTF8}{gbsn}\textcolor{red}{甲亢患者教育：}\textcolor{blue}{1. 哺乳期伴甲状腺功能亢进症患者首选丙硫氧嘧啶。}\end{CJK} \textcolor{red}{Education for patients with hyperthyroidism: } \textcolor{blue}{1. Propylthiouracil is the first choice for patients with hyperthyroidism during lactation.}\\
        \hline
    \end{tabular}
    }
    \caption{Examples of generated explanations. Unrelated content is shown in \textcolor{red}{red}, and key phrases used for illustration are shown in \textcolor{blue}{blue}.}
    \label{tab:case}
    
\end{table*}
\subsection{Case Study}
Table \ref{tab:case} presents the generated explanations produced by different methods from the CME test set. Our method uses the pre-trained generative model to produce a reasonable and concise explanation. Specifically, we show that both the generated explanation and the predicted labels are correct in the example; our method can learn the treatment of a specific drug under a specific condition. Compared to other models, our method can generate explanations that are well correlated with labels. It provides more accurate information than a simple negative wording. We also compare the generation explanations of GPT2-style models pre-trained by~\cite{wei2019nezha} (i.e., NEZHA-GEN~\footnote{\url{https://github.com/huawei-noah/Pretrained-Language-Model}}).

\section{Discussion}
\label{sec:discussion}
Humans learn and understand the world partly through the integration of new information into existing knowledge. Generating explanations to oneself (i.e., self-explaining) facilitates the integration process. Self-explanation has been shown to improve the acquisition of problem-solving skills from cognitive science research~\cite{vanlehn1992model,chi1994eliciting}. 
We posit that such self-explanations are also important and necessary for natural language understanding models.

Our findings provide quantitative evidence to:
(1) Joint learning with explanation generation improves model performance but does not lead to statistically significant improvements. It suggests that models still struggle to learn to generate explanations sufficiently; 
(2) There are two possible setups to output generative explanations: generating an explanation for each option (label-specific), or only generating one explanation. We show that it is sufficient to generate only one explanation that explains why the correct answer was chosen for each question; 
(3) A cause that explains in a more informative and fluent manner may be considered as a reasonable-sounding explanation to humans.

As one limitation of this work, human-annotated explanations could be biased and correlated with superficial clues, and how the machine can help us correct them instead of learning to replicate them could be another interesting topic. Our goal of interpretability study is not totally about trust or confidence in PLM-generated text, while we focus the ability for professionals to understand the justifications for model decisions as one step.

\section{Conclusion}
\label{sec:conclusion}
In this paper, we propose a \textit{self-explaining} method (\approach{}) to make the prediction and generate its corresponding explanation simultaneously. We evaluate the proposed \approach{} on three tasks. Experiment results show that our approach yields improvements in these tasks. Human evaluations demonstrate that generated explanations are well correlated with the predicted labels. Moreover, we find that the model improves the performance when trained with samples of no natural language explanations. 
In the future, we will adopt more evaluation criteria on the explanations in order to analyze the sufficiency and comprehensiveness.



\bibliography{acl2021}{}
\bibliographystyle{unsrtnat}

\end{document}